\pdfoutput=1

\documentclass[11pt]{article}

\usepackage[]{acl}

\usepackage{times}
\usepackage{latexsym}

\usepackage[T1]{fontenc}

\usepackage[utf8]{inputenc}

\usepackage{microtype}

\usepackage{inconsolata}
\usepackage{booktabs}
\usepackage{multirow}
\usepackage{hyperref}
\usepackage{rotating}
\usepackage{enumitem}
\usepackage{tablefootnote}
\usepackage{pifont}

\usepackage{pdfpages}
\usepackage{longtable}
\usepackage{setspace}
\usepackage{etoolbox,xspace}
\usepackage{graphicx}
\usepackage{placeins}

\newcommand*{\gemmanine}{Gemma-2-9b\xspace}
\newcommand*{\llamathree}{Meta-LLaMA-3-8B\xspace}
\newcommand*{\llamas}{Meta-LLaMA-3-70B\xspace}
\newcommand*{\gemmats}{Gemma-2-27b\xspace}
\newcommand*{\aya}{Aya-101\xspace}
\newcommand*{\llamax}{LLaMAX3-8B-Alpaca\xspace}
\newcommand*{\gptf}{Gpt-4o\xspace}
\newcommand*{\proverbeval}{\textbf{ProverbEval}\xspace}

\usepackage[toc,page]{appendix}

\usepackage{caption}
\usepackage{enumitem}
\usepackage{graphicx}
\usepackage{amssymb}
\usepackage{xspace}
\usepackage[ethiop,main=english]{babel}
\usepackage{inconsolata}
\usepackage[most]{tcolorbox}
\usepackage{graphicx}

\usepackage{pifont}
\usepackage{float}

\newcommand\blfootnote[1]{%
  \begingroup
  \renewcommand\thefootnote{}\footnote{#1}%
  \addtocounter{footnote}{-1}%
  \endgroup
}

\title{ProverbEval: Exploring LLM Evaluation Challenges for Low-resource Language Understanding}

\author{\normalsize Israel Abebe Azime$^{1,\ast ,\dagger}$,  Atnafu Lambebo Tonja$^{2,3, \ast,\dagger}$, Tadesse Destaw Belay$^{4,\dagger}$,   \\ 
\textbf{\normalsize Yonas Chanie$^{5,\dagger}$, Bontu Fufa Balcha $^{6,\dagger}$, Negasi Haile Abadi $^{7,\dagger}$, Henok Biadglign Ademtew$^{8,\dagger}$, } \\
\textbf{\normalsize  Mulubrhan Abebe Nerea$^{9,\dagger}$, Debela Desalegn Yadeta $^{6}$, Derartu Dagne Geremew$^{10,\dagger}$}, \\
\textbf{\normalsize    Assefa Atsbiha tesfau$^{7,\dagger}$, Philipp Slusallek$^{1}$, Thamar Solorio$^{2}$,  Dietrich Klakow$^{1}$}, \\\\
\footnotesize
 $^\dagger$ Ethio NLP, $^1$ Saarland University, $^2$ MBZUAI, $^3$ Lelapa AI,  $^4$ Instituto Politécnico Nacional,
\\ 
 \footnotesize
   $^5$Pindo, $^6$ AAIT, $^7$Lesan AI, $^8$EAII, $^9$University West, $^{10}$Haramaya University,
\\}

\begin{document}
\maketitle
\blfootnote{$^\ast$ Equal Contribution.}
\begin{abstract}
With the rapid development of evaluation datasets to assess LLMs understanding across a wide range of subjects and domains, identifying a suitable language understanding benchmark has become increasingly challenging. In this work, we explore LLM evaluation challenges for low-resource language understanding and introduce \proverbeval, LLM evaluation benchmark for low-resource languages, focusing on low-resource language understanding in culture-specific scenarios. We benchmark various LLMs and explore factors that create variability in the benchmarking process.  We observed performance variances of up to 50\%, depending on the order in which answer choices were presented in multiple-choice tasks. Native language proverb descriptions significantly improve tasks such as proverb generation, contributing to improved outcomes. Additionally, monolingual evaluations consistently outperformed their cross-lingual counterparts in generation tasks.
We argue that special attention must be given to the order of choices, the choice of prompt language, task variability, and generation tasks when creating LLM evaluation benchmarks\footnote{Evaluation data available at \url{https://huggingface.co/datasets/israel/ProverbEval} evaluation code \url{https://github.com/EthioNLP/EthioProverbEval}}.
  
\end{abstract}

\section{Introduction}
Large language models (LLMs) evaluation is gaining increasing attention as these models are typically trained on general-domain datasets while demonstrating notable performance on tasks out of their training domains \cite{mosbach2023few}. The creation of evaluation datasets helps to identify the capabilities of LLMs, pinpoint shortcomings, and establish a measurable path for improvement. Based on \citet{chang2024survey}, LLM evaluation addresses questions such as what to evaluate (subjects and topics), where to evaluate (selecting appropriate datasets), and how to evaluate (the evaluation process). 

To improve LLMs' capabilities and effectively assess their performance, researchers are creating benchmark datasets using a diverse range of domains and languages. This inclusive methodology allows for a more comprehensive evaluation of LLMs' performance across various domains and languages. Popular benchmark datasets like MMLU \cite{hendrycks2020measuring} and MEGAVERSE \cite{ahuja2023megaverse} cover a wide range of extensive world knowledge tasks and subjects.

To create evaluation benchmarks that are multilingual, researchers \citet{koto2024arabicmmlu,li2023cmmlu,son2024kmmlu} introduced benchmark datasets for different languages by translating a subset of the MMLU dataset. Beyond research efforts, translating existing benchmarks into different languages is an effective strategy to evaluate the multilingual capabilities of closed-source LLMs. 
These benchmarks evaluate multilingual understanding of models by presenting a range of extensive world knowledge tasks in the language of interest. While combining different subjects in a benchmark dataset may seem beneficial, it does not always provide a clear picture of the model’s shortcomings. For example, using MMLU in different languages tests language and subject understanding simultaneously \cite{hendrycks2020measuring}. There should be evaluation benchmarks that disentangle language understanding and specific subject knowledge.


Language understanding of LLM can be measured in numerous ways, and it is crucial to introduce benchmarks that evaluate complex text comprehension while considering each language's specific linguistic, cultural, and contextual nuances. Creating benchmarks tailored to individual languages' unique values and customs is essential for ensuring comprehensive and accurate evaluations of language models \cite{liu2023multilingual}.




\begin{quote}
    \textit{``If culture was a house, then language was the key to the front door, to all the rooms inside.''} — {\footnotesize Khaled Hosseini, Afghan-born American novelist and physician}
    \end{quote}

Language plays a vital role in shaping and preserving cultural identity \cite{wang-etal-2024-countries}. It serves as a medium for not only communication but also for the transmission of traditions, values, and beliefs from one generation to another. Through language, individuals can express their emotions, share their stories, and form deep connections with others. With approximately 7000 spoken languages across the globe, each language reflects the unique history, customs, and perspectives of the community that speaks it \cite{zheng-etal-2024}. 

A proverb is a short, well-known pithy saying, stating a general truth or a piece of advice. Proverbs are like windows into a culture, offering brief but powerful insights into how people think and live. They carry lessons, reflect shared values, and communicate wisdom passed down through generations. They are a rich manifestation of a society's values, beliefs, and worldview and serve valuable didactic and communicative purposes \cite{LOMOTEY202186}. For example, the English proverb \textit{The apple does not fall far from the tree} — means a child grows up to resemble his/her parents. While a plain version of this proverb exists in many cultures, it is expressed differently in different languages and cultures \citep{liu2023multilingual}. For instance, the above proverb might be equivalent in meaning to an Ethiopian Proverb “\selectlanguage{ethiop}le^ge 'abAtune 'ayebe 'aguAtune yemaselAle”\selectlanguage{english} — literally meaning the son resembles his father, the cheese its milk. 





In this paper, we introduce \proverbeval: LLM evaluation dataset with three distinct tasks based on cultural proverbs for 4 Ethiopian languages and English.
The contributions of this work are as follows:
\begin{itemize}
    \item Introduce \proverbeval, a comprehensive LLM evaluation dataset comprising three distinct tasks, derived from cultural proverbs in four Ethiopian languages and English.

    \item Explore zero-shot performances of a wide range of LLMs on monolingual and cross-lingual language understanding abilities for low-resource languages.
    
    \item Explore LLM evaluation challenges for low-resource language understanding. 

    
    
\end{itemize}

\begin{figure*}[!h]
    \centering
   \includegraphics[width=\linewidth]{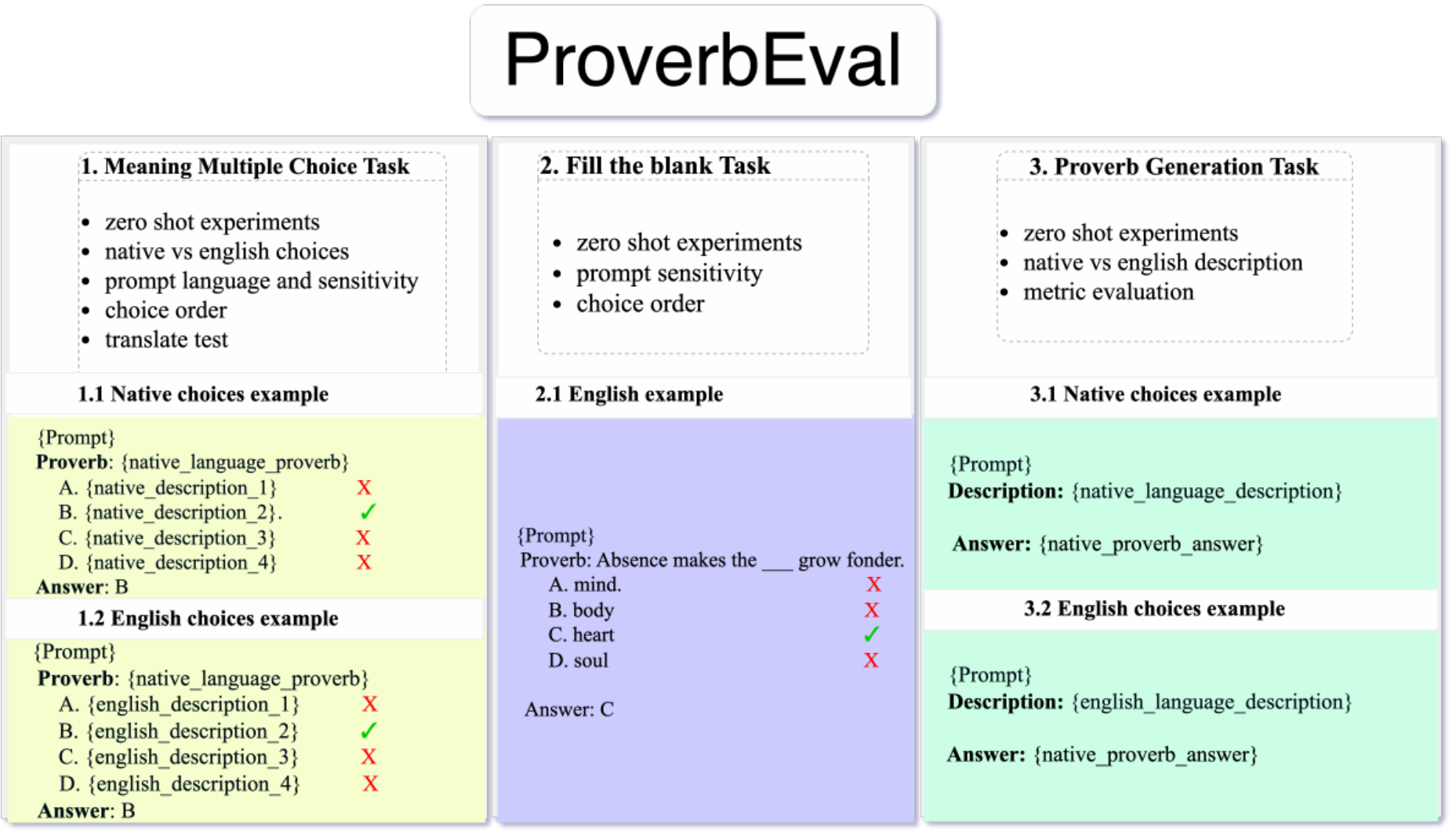}
    \caption{Detailed overview of \proverbeval, which consists of three distinct tasks. Native languages include those included in Table \ref{tab:data-dist}.
Detailed prompt descriptions can be found in Appendix \ref{app:prompt-details}. }
    \label{fig:task-spec}
\end{figure*}


\section{Related Work}
\label{related-work}


Significant efforts have been made to include diverse languages in the development of multilingual language models \cite{conneau-etal-2020-unsupervised,xue-etal-2021-mt5}. \citet{rust-etal-2021-good} conducted a comparison between multilingual and monolingual language models, employing metrics such as subword fertility. Subword fertility, defined as the ratio of subtokens to total tokens, has been shown to have a direct correlation with model performance across languages, illustrating the impact of tokenization on multilingual language model efficacy. Apart from architecture-based evaluation, multilingual benchmarks help us to track the progress toward multilingualism.

Current evaluation benchmarks prioritize multiple-choice questions due to the relative ease of automatic scoring, as opposed to open-ended question benchmarks that demand significant human involvement \cite{son2024kmmlu,wang2024mmlu}. For example, MMLU-Pro \cite{wang2024mmlu} places a strong emphasis on prompt variations and their influence on large language model (LLM) performance. 



Cultural significance of LLM benchmarks is crucial factor to consider as part of language understanding. To incorporate cultures into benchmarks, \citet{myung2024blend} introduced  BLEnD, which covers 16 countries and 13 languages to prepare datasets that have tests of significance for users in their region. Additionally \citet{liu2023multilingual} shows proverbs can be used to assess LLMs cultural understanding in several languages and introduces \textbf{MAPS} (Multicultural Proverbs and Sayings) dataset based on proverbs and sayings to evaluate LLMs multilingual and cultural understanding ability. Our work adopts the same motivation to use proverbs and expands it to different languages and task types.

\section{Methodology}
\label{Methodology}

\subsection{Languages Covered}
We create \proverbeval benchmark dataset for four low-resource languages along with English to evaluate the cross-lingual capability of LLMs. From these languages, three languages were written in Ethiopic script: Amharic, Tigrinya, and Ge'ez, and two languages in Latin script: English and Afaan Oromo. We begin with these languages due to the availability of native speaker access to construct the dataset.


\begin{table}[hbt!]
    \centering
    \resizebox{\linewidth}{!}{%
    \begin{tabular}{lcccc}
    \hline
    \textbf{Language}&\textbf{\# Task 1}  & \textbf{\# Task 2} &\textbf{\# Task 3} \\
    \hline
    \texttt{Amharic}  & 483 & 494 & 484\\
    \texttt{Afaan Oromo}  & 502 & 493 & 502 \\
    \texttt{Tigrinya}  & 380 & 503 & 380\\ 
    \texttt{Ge'ez}  & 434 & 429 & 434\\
    \texttt{English}  & 437 & 462 & 437\\
    \hline
    \end{tabular}
    }
    \caption{\proverbeval languages and data sizes. All numbers show the test set data size that was prepared. }
    \label{tab:data-dist}
\end{table}

\subsection{Data Collection}
Proverbs belong to the public domain and are widely regarded as shared cultural expressions. Their public domain status allows us to freely collect and utilize these resources without licensing restrictions.  We collect proverbs from books, online sources, and the common knowledge of volunteer annotators. Our data collection focuses on collecting proverbs, writing detailed explanations in native and English languages, and verifying the correctness of the collected data. 

The data collection was carried out by volunteers who are contributing to this research as co-authors.
Data collectors utilized existing machine translation (MT) systems to verify and supplement any vocabulary gaps they encountered while writing proverbs in English after completing explanations in native languages. 

As shown in Table \ref{tab:data-dist}, we focused on collecting only the test sets for all tasks. Additionally, we included five items that can serve as few-shot examples for \textit{Task 2: Fill in the Blank}.

\paragraph{Biases in Proverbs:}
The compact and metaphorical language in proverbs is intriguing, but it can also serve as a tool to reinforce gender stereotypes and racial inequalities. In this work, we gave special attention to proverbs that reflect these values and removed all instances of such proverbs.

\subsection{Tasks}
\proverbeval benchmark contains three main tasks: multiple choice, fill-the-blank, and generation tasks with various evaluation settings. 
\paragraph{Task 1: Meaning Multiple Choice}
In meaning-based multiple-choice tasks, we aim to assess the model’s language understanding capabilities by asking the model to select the option with the most similar meaning. For each proverb, four options are provided, each with a detailed explanation of its possible meaning, with only one being correct.

\textbf{Native vs English choices} -- One of the factors we are currently exploring in our experiment is the selection of language used in the multiple-choice options. This exploration will help us access the cross-lingual capability in addition to the monolingual capability of models where the proverb is given, and the model has to choose a sentence that closely resamples it. Figure \ref{fig:task-spec} explains details of task variations. Due to the extremely low resource availability for \textbf{\texttt{Ge’ez}, proverb descriptions} are carried out using Amharic, a closely related language.

\paragraph{Task 2: Fill in the Blank}

The fill-in-the-blank task is designed to evaluate: 
the ability of the model to recall proverbs despite containing unconventional word order. For example, the proverb "\textit{Don’t let the cat out of the \_\_\_} " commonly should be followed by \textit{house} rather than \textit{bag} if we do not have an understanding of that specific proverb, as cats are more commonly associated with houses rather than bags.  In this task, we will assess how well the LLMs understand the common proverb.

\paragraph{Task 3: Generation}
The ability to determine which proverb best aligns with a particular meaning or situation serves as a way to assess and measure a model's understanding of language. In order to evaluate this, we designed a generation task in which a detailed description of the proverb is provided, and the model is required to select the most appropriate proverb that aligns with the description given.
We chose this approach for easier evaluation, though the dataset could also be used for tasks involving generating descriptions based on a given proverb. 

\paragraph{Native and English descriptions} -
For this task, we utilized both \textit{native} language and \textit{English} descriptions. Descriptions provided in English with the expectation of receiving a native proverb allowed us to evaluate cross-lingual capabilities. Conversely, descriptions given in the native language with the expectation of a corresponding native proverb enabled us to assess monolingual comprehension.

\begin{figure}[!ht]
    \centering
    \includegraphics[width=\linewidth]{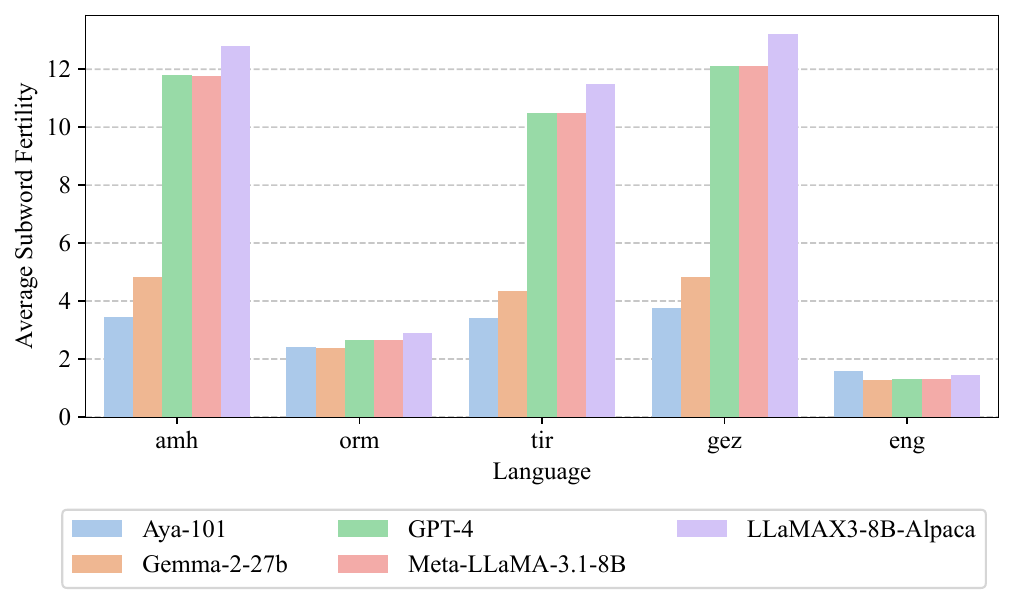}
    \caption{Subword fertility of proverbs for each model’s tokenizer in our study. Models that share the same tokenizers are grouped together. Lower values indicate better performance, as they reflect that words are not being excessively split on average. }
    \label{fig:subword-fertility}
\end{figure}

\begin{table*}[hbt!]
    \centering
    \resizebox{0.95\linewidth}{!}{%
    \begin{tabular}{l|ll|ll|ll|ll|l|lll}
        \toprule
        \multirow{2}{7em}{\textbf{Model Name} \\ \textit{prompt language}}  & \multicolumn{2}{|c|}{\textbf{Amharic}} & \multicolumn{2}{|c|}{\textbf{Afaan Oromo}} & \multicolumn{2}{|c|}{\textbf{Tigrinya}} & \multicolumn{2}{|c|}{\textbf{Ge'ez} } & \multirow{2}{3em}{\textbf{English}} & \multicolumn{3}{|c}{\textbf{average}} \\ 
        \textbf{} & \textit{native} & \textit{english} & \textit{native} & \textit{english} & \textit{native} & \textit{english} & \textit{native} & \textit{english} &  & \textit{native} & \textit{english} & \textit{all} \\ \midrule
        \textbf{\llamathree} & ~ & ~ & ~ & ~ & ~ & ~ & ~ & ~ & ~ & ~ & ~ & ~ \\
        \textit{English prompts} & 24.72 & 24.98 & 32.54 & 25.37 & 26.93 & 29.83 & 30.11 & 29.27 & 49.43 & 28.58 & 27.36 & 30.35 \\ 
        \textit{Native Prompts} & 31.54 & 26.43 & 26.23 & 24.97 & 27.11 & 25.09 & 26.42 & 24.19 & ~ & 27.83 & 25.17 & 26.50 \\ \hline
        \textbf{\gemmanine} & ~ & ~ & ~ & ~ & ~ & ~ & ~ & ~ & ~ & ~ & ~ & ~ \\
        \textit{English prompts} & 31.06 & 30.85 & 29.22 & 26.43 & 29.82 & 30.88 & 38.1 & 45.93 & 63.31 & 32.05 & 33.52 & 36.18 \\ 
        \textit{Native Prompts} & 29.41 & 34.77 & 25.30 & 26.69 & 28.07 & 26.93 & 26.74 & 26.04 & ~ & 27.38 & 29.46 & 28.27 \\ \hline
        \textbf{\gemmats} & ~ & ~ & ~ & ~ & ~ & ~ & ~ & ~ & ~ & ~ & ~ & ~ \\
        \textit{English prompts} & 35.06 & 36.3 & 34.99 & 27.69 & 32.39 & 33.95 & 41.86 & 42.71 & 68.12 & 36.08 & 35.16 & 39.23 \\ 
        \textit{Native Prompts} & 38.36 & 39.54 & 25.57 & 26.89 & 25.17 & 25.53 & 27.65 & 25.04 & ~ & 29.19 & 30.65 & 29.82 \\ \hline
        \textbf{\llamas} & ~ & ~ & ~ & ~ & ~ & ~ & ~ & ~ & ~ & ~ & ~ & ~ \\
        \textit{English prompts} & 41.67 & 37.61 & 32.67 & 27.96 & 36.49 & 30.96 & 55.07 & 47.62 & 71.70 & 41.48 & 36.04 & 42.42 \\ 
        \textit{Native Prompts} & 26.24 & 27.19 & 27.09 & 25.76 & 26.67 & 27.11 & 26.27 & 25.20 & ~ & 26.57 & 26.32 & 26.44 \\ \hline 
        \textbf{\llamax} & ~ & ~ & ~ & ~ & ~ & ~ & ~ & ~ & ~ & ~ & ~ & ~ \\
        \textit{English prompts} & 28.99 & 25.38 & 31.94 & 25.77 & 29.21 & 28.25 & 35.02 & 30.72 & 42.71 & 31.29 & 27.53 & 30.89 \\ 
        \textit{Native Prompts} & 30.09 & 26.15 & 26.16 & 26.69 & 27.17 & 25.97 & 26.42 & 25.12 & ~ & 27.46 & 25.98 & 26.72 \\ \hline
        \textbf{\aya} & ~ & ~ & ~ & ~ & ~ & ~ & ~ & ~ & ~ & ~ & ~ & ~ \\
        \textit{English prompts} & \textbf{48.21} & 52.38 & \textbf{49.40} & 32.80 & 42.19 & \textbf{55.09} & \textbf{75.34} & \textbf{82.49} & 77.42 & \textbf{53.79} & \textbf{55.69} & \textbf{57.26} \\ 
        \textit{Native Prompts} & 50.96 & \textbf{55.21} & 41.97 & 28.82 & \textbf{49.74} & 32.72 & 45.00 & 48.69 & ~ & 46.92 & 41.36 & 44.14 \\ \hline 
          \textbf{\gptf} & ~ & ~ & ~ & ~ & ~ & ~ & ~ & ~ & ~ & ~ & ~ & ~ \\
        \textit{English prompts} & 40.19 & 46.24 & 49.01 & 50.80 & 32.37 & 35.00 & 24.20 & 59.29 &  \textbf{89.97}& 36.44 & 47.83 & 47.45 \\ 
        \textit{Native Prompts} & 44.42 & 48.93 & 27.22 & \textbf{55.31}& 24.82 & 7.54 & 0.08 & 1.15 & ~ &24.13 &28.23  & 26.18  \\ 
        
        \bottomrule
       
    \end{tabular}   }
    
    \caption{Zero-shot scores of Task 1 (\textit{ meaning multiple choice task}) across all models for English and native prompts for choosing from \textbf{\textit{native}} choices and \textbf{\textit{english}} choices. All scores are average of 3 distinct prompts. prompt details in Appendix \ref{app:prompt-details} and detailed results in \ref{app:Prompt Sensitivity}.}
    \label{nativevsenglish}
\end{table*}

\section{Experimental Setup and setting}
\label{Experiments}

\subsection{Model Selection}
Given the wide range of available model options, we established criteria to guide model selection. The models chosen for this experiment were based on the following key factors: (1) different models in terms of the number of parameter size, (2) closed-source versus open-source models, (3) multilingual models versus general-purpose models, and (4) instructed models versus base models. 

In this experiment, we did not include open-source instruction-finetuned models due to the difficulty in accessing the specific instruction-finetuning data used for their training. Instead, we utilized \llamax, which is finetuned on the Alpaca dataset, and \aya, which incorporates a combination of various task-oriented and generative datasets.
For large models, we include \llamas \cite{dubey2024llama}and \gemmats \cite{team2024gemma}; for average size models, we included \llamathree \cite{dubey2024llama} and \gemmanine \cite{team2024gemma}; for multilingual models, we include \aya \cite{ustun-etal-2024-aya} and \llamax \cite{lu2024llamax}; finally, we included \gptf \cite{achiam2023gpt} from closed source models. From the model list, we select \aya model since it is mT5 based model used to compare with decoder-only models.

\subsection{Evaluation}
We used ElutherAI's open-source Language Model Evaluation Harness (lm-eval) framework \cite{eval-harness} to evaluate the models. The library supports evaluation strategies, including log-likelihood, generation, and perplexity, using \textit{YAML} to configure and manage the evaluations. 
We used log-likelihood and generation for open-source models for multiple-choice and fill-the-blank tasks. In the multiple-choice task and fill-the-blank, each option is appended to the corresponding question and prompt, and the log-likelihood score is subsequently computed for evaluation. Finally, the accuracy score is reported to be the highest selected option. 


For Generation tasks, we heavily rely on ChrF \cite{popovic-2015-ChrF} scores but included BLEU and translation edit rate (ter) \cite{snover-etal-2006-study} scores in the Appendix \ref{app:generation-results}. 

For \gptf evaluation, we used the generate\_until output type for all tasks since it does not support log-likelihood. We wrote a verbalizer to extract answers from generated answers and calculate accuracy scores for all tasks except for generation.  
\subsection{Experiments}
\subsubsection{Zero-shot evaluation of the models}

In our first experiment, we performed a comprehensive zero-shot evaluation on all tasks. This involved rigorously testing the LLMs language understanding capabilities by subjecting it to our carefully curated test set.

\subsubsection{Key Factors Influencing Zero-Shot Performance}

Most LLM evaluation benchmarks rely on multiple-choice tasks due to the ease of evaluation. Compared to generative tasks, multiple choice tasks are simpler to assess using automatic metrics, as they eliminate the possibility that the model provides a correct answer in a different form from the ground truth \cite{zhang2024multiple}. This approach ensures consistency in the evaluation and avoids ambiguity when assessing the model's performance.

In this work, in addition to introducing proverb-based tasks, we are interested in exploring the reliability of multiple-choice evaluations. To answer this question, we explored the following factors.

\paragraph{Prompting language} is one factor that affects the performance of the model. Models can be sensitive to different prompts and prompts given in several languages \cite{zhang2023prompting}.  To evaluate the effect, we tested three English prompts to assess model performance with diverse English inputs and three native prompts for each language to assess performance with instructions in the respective native language.


\paragraph{Order of choices } affects the performance of the models in multiple-choice tasks  \cite{zheng2023large,pezeshkpour2023large}. To evaluate the effect of this problem in a low-resource scenario, we compared the average of three random shuffle performances of the models to correct answers appearing first (all "A") or last choice (all "D").

\begin{table*}[!ht]
    \centering
    \resizebox{0.85\linewidth}{!}{%
    \begin{tabular}{l|ll|ll|ll|ll|l|lll}
        \toprule
        \multirow{2}{7em}{\textbf{Model Name} \\ \textit{shuffling strategy}} & \multicolumn{2}{|c|}{\textbf{Amharic} } & \multicolumn{2}{|c|}{\textbf{Afaan Oromo} } & \multicolumn{2}{|c|}{\textbf{Tigrinya} } & \multicolumn{2}{|c|}{\textbf{Ge'ez} } & \textbf{English} & \multicolumn{3}{|c}{\textbf{Average}} \\ 
        \textbf{} & \textit{native} & \textit{english}  & \textit{native} & \textit{english} & \textit{native} & \textit{english}  & \textit{native} & \textit{english}  &  & \textit{native} & \textit{english}  & \textit{all} \\ \midrule
        \textbf{\llamathree} & ~ & ~ & ~ & ~ & ~ & ~ & ~ & ~ & ~ & ~ & ~ & ~ \\ 
        \textit{3 random shuffle} & 26.86 & 26.98 & 31.54 & 25.77 & 29.30 & 26.05 & 30.34 & 27.67 & 50.73 & 29.51 & 26.62 & 30.58 \\ 
        \textit{all option A} & 58.88 & 73.91 & 69.92 & 80.08 & 55.79 & 80.26 & 69.12 & 77.63 & 89.47 & 63.43 & 77.97 & 72.78 \\ 
        \textit{all option D} & 7.23 & 9.94 & 16.73 & 4.98 & 3.95 & 2.89 & 4.38 & 4.47 & 23.57 & 8.07 & 5.57 & 8.68 \\ \hline
        \textbf{\gemmanine} & ~ & ~ & ~ & ~ & ~ & ~ & ~ & ~ & ~ & ~ & ~ & ~ \\ 
        \textit{3 random shuffle } & 29.68 & 33.89 & 31.54 & 28.89 & 29.47 & 27.46 & 40.4 & 27.37 & 64.23 & 32.77 & 29.4 & 34.77 \\ 
        \textit{all option A} & 63.02 & 81.16 & 69.12 & 88.65 & 50.53 & 87.63 & 73.27 & 86.05 & 90.85 & 63.99 & 85.87 & 76.70 \\ 
        \textit{all option D} & 22.11 & 13.66 & 24.10 & 4.38 & 24.47 & 5.00 & 31.34 & 5.05 & 50.80 & 25.51 & 7.02 & 20.10 \\ \hline
        \textbf{\gemmats} & ~ & ~ & ~ & ~ & ~ & ~ & ~ & ~ & ~ & ~ & ~ & ~ \\ 
        \textit{3 random shuffle } & 34.64 & 34.57 & 36.25 & 29.02 & 28.16 & 29.91 & 40.02 & 29.82 & 66.13 & 34.77 & 30.83 & 36.50 \\ 
        \textit{all option A} & 65.29 & 69.57 & 62.35 & 74.50 & 55.00 & 73.68 & 72.81 & 75.00 & 90.39 & 63.86 & 73.19 & 70.95 \\ 
        \textit{all option D} & 19.21 & 20.7 & 25.7 & 9.76 & 18.16 & 7.11 & 27.19 & 8.68 & 54.23 & 22.57 & 11.56 & 21.19 \\ \hline
        \textbf{\llamas} & ~ & ~ & ~ & ~ & ~ & ~ & ~ & ~ & ~ & ~ & ~ & ~ \\ 
        \textit{3 random shuffle } & 41.94 & 40.30 & 35.13 & 31.51 & 38.16 & 30.18 & 61.44 & 30.26 & 74.52 & 44.17 & 33.06 & 42.60 \\ 
        \textit{all option A} & 50.62 & 53.21 & 56.77 & 50.40 & 41.84 & 58.68 & 71.66 & 58.68 & 79.86 & 55.22 & 55.24 & 57.97 \\ 
        \textit{all option D} & 28.51 & 21.33 & 21.71 & 14.74 & 20.53 & 9.47 & 42.63 & 10.26 & 71.85 & 28.35 & 13.95 & 26.78 \\ \hline
        \textbf{\llamax} & ~ & ~ & ~ & ~ & ~ & ~ & ~ & ~ & ~ & ~ & ~ & ~ \\ 
        \textit{3 random shuffle } & 33.95 & 25.33 & 31.48 & 26.29 & 30.79 & 27.63 & 33.18 & 27.19 & 39.51 & 32.35 & 26.61 & 30.59 \\ 
        \textit{all option A} & 36.98 & 64.39 & 47.81 & 72.11 & 25.79 & 79.47 & 39.86 & 80.53 & 75.97 & 37.61 & 74.13 & 58.10 \\ 
        \textit{all option D} & 34.5 & 10.56 & 21.91 & 4.38 & 37.63 & 4.47 & 30.65 & 4.21 & 24.03 & 31.17 & 5.91 & 19.15 \\ \hline
        \textbf{\aya} & ~ & ~ & ~ & ~ & ~ & ~ & ~ & ~ & ~ & ~ & ~ & ~ \\ 
        \textit{3 random shuffle } & 51.24 & 54.98 & 51.06 & 32.8 & 43.16 & 55.35 & 78.88 & 55.88 & 80.78 & 56.09 & 49.75 & 56.01 \\ 
        \textit{all option A} & 61.98 & 67.91 & 54.58 & 44.62 & 51.32 & 67.63 & 85.71 & 70.26 & 82.38 & 63.4 & 62.61 & 65.15 \\ 
        \textit{all option D} & 57.23 & 56.73 & 51.99 & 31.27 & 50.53 & 50.53 & 81.80 & 49.21 & 80.78 & 60.39 & 46.94 & 56.67 \\ \hline
        \textbf{\gptf} & ~ & ~ & ~ & ~ & ~ & ~ & ~ & ~ & ~ & ~ & ~ & ~ \\ 
        \textit{3 random shuffle } & 59.51 & 52.86 & 78.75 & 75.43 & 43.33 & 26.05 & 51.92 & 22.79 & 86.96 & 65.66 &	66.41 &	69.75 \\ 
        \textit{all option A} & 52.27 & 43.48 & 80.48 & 76.10 & 43.16 & 23.42 & 91.94 & 23.68 & 99.54 & 68.13 &	59.71 & 67.88 \\ 
        \textit{all option D} & 50.00 & 45.55 & 72.71 & 77.89 & 35.00 & 11.32 & 77.42 & 13.42 & 99.08 & 59.35 &	53.51 &	61.17 \\ 
        \bottomrule
    \end{tabular}
    }
    \caption{Zero-shot accuracy scores of Task 1 (\textit{ meaning multiple choice task}) across all models for \textbf{\textit{native choices}} and \textbf{\textit{English choices}}.}
    \label{tab:shuffle_stuff}
\end{table*}

\paragraph{Few-shot Experiments}
For \textit{task 2: proverb fill the blank task}, we explored if introducing examples can improve the performance of the models using our validation set. 

\paragraph{Effect of Translation}
Cross-linguistic translation of proverbs is challenging because these expressions often carry culturally specific meanings that may not have direct equivalents in other languages. When proverbs are translated, the nuances and cultural significance can be lost, making it difficult for non-native speakers to fully understand the intended message. Our analysis of closed-source models indicates that LLMs mitigate their lack of language understanding by translating questions from low-resource languages to English and conducting reasoning in English.
We translated our proverbs and compared them with the native ones to see if our task is easily solvable by translating, as shown in Table \ref{table:translate-test}.




\section{Results and Analysis}
\label{Results}

\subsection{Proverb Multiple Choice}
\paragraph{Does model size significantly improve performance for low-resource languages?} 

In Table \ref{nativevsenglish}, the result indicates that the size of the open-source base models has a notable impact on the prompt that is being used. Generally, the bigger the models, the better, but \gemmats takes the lead in \textit{native} prompt, and \llamas takes the lead in \textit{english} prompt. This directly correlates with Figure \ref{fig:subword-fertility} that the model with the lowest sub-word fertility is the better and Gemma models are better multilingual models. This is more reflected in Gemma models having better performance in \textit{native} choices than \textit{English} choices. We can conclude that a better tokenizer (lower monolingual fertility) is very important in monolingual evaluation compared to cross-lingual evaluation. 



\begin{figure*}[h!]
    \centering
    \includegraphics[width=\linewidth]{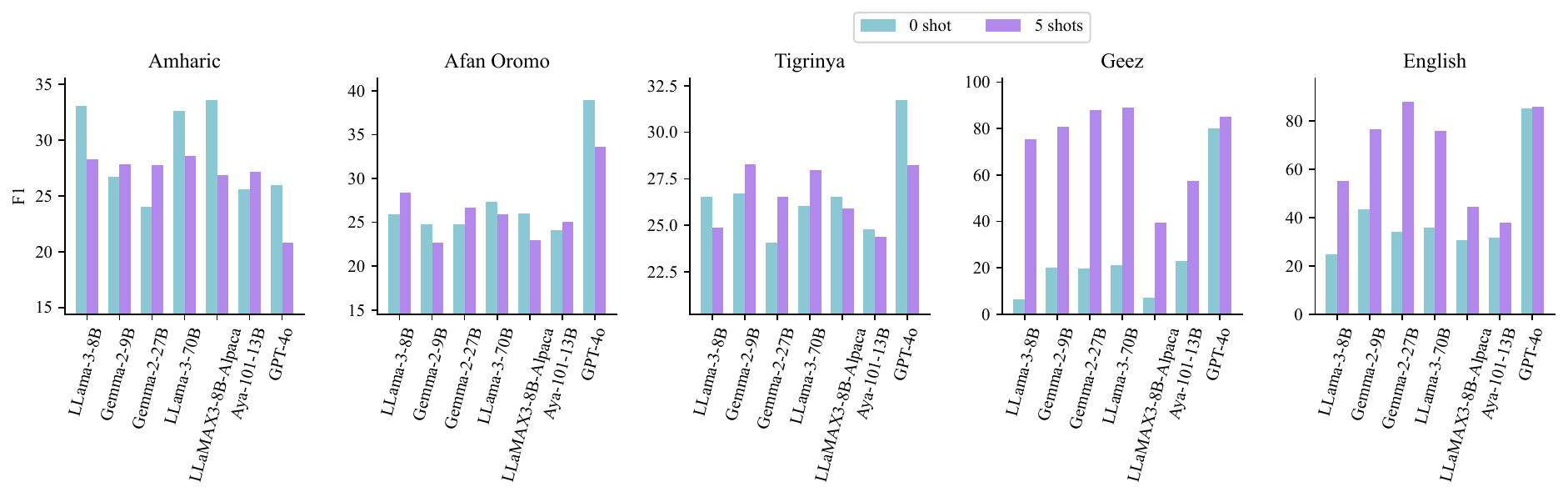}
    \caption{Average accuracy of fill-the-blank results (0 and 5 shots). Zero-shot and five-shot results are an average of three random shuffles using English prompt.}
    \label{fig:zero-shot-all}
\end{figure*}

\paragraph{Does the choice of language in the prompt affect performance for low-resource languages?} For tasks using native prompts, show lower results compared to using English prompts for non-English languages. Using in-language prompt results min 0 and max ±3 differences between \textit{native} and \textit{english} multiple choice in task 1.

As seen in Table \ref{nativevsenglish}, only the biggest or multilingual fine-tuned models show promising results in \textit{meaning multiple choice task}. The results in English also show that the task is answerable with a specific focus on languages, and this dataset will be an important resource to identify whether LLMs will achieve meaningful reasoning ability in low-resource languages. Additionally, as we can see from the table, \gptf shows better results when choices are given in English than in native languages.

\paragraph{Are LLMs sensitive to choice order in low-resource languages?}
 As shown in Table \ref{tab:shuffle_stuff}, smaller models show a difference of accuracy close to 30\% and 50\% when the answers are provided in the first choice. This number decreases significantly when using larger models or when testing models that pass through supervised fine-tuning. \aya model shows resistance to this disturbance probably because of the training data containing several tasks, whereas the \gptf model shows persistent results regardless of choice order.  
Looking at \textit{native} and \textit{English} choices for all prompts, we can clearly see that choice order affects cross-lingual tasks more than monolingual tasks. 

%

\begin{table*}[!ht]
    \centering
    \resizebox{\linewidth}{!}{%
    \begin{tabular}{l|ll|ll|ll|ll|l|lll}
    \hline
    \toprule
    \multirow{2}{8em}{\textbf{Model Name} } & \multicolumn{2}{|c|}{\textbf{Amharic} } & \multicolumn{2}{|c|}{\textbf{Afaan Oromo} } & \multicolumn{2}{|c|}{\textbf{Tigrinya} } & \multicolumn{2}{|c|}{\textbf{Ge'ez} } & \textbf{English} & \multicolumn{3}{|c}{\textbf{Average}} \\ 
    \textbf{} & \textit{native} & \textit{english}  & \textit{native} & \textit{english} & \textit{native} & \textit{english}  & \textit{native} & \textit{english}  &  & \textit{native} & \textit{english}  & \textit{all} \\ \midrule
        \llamathree & 1.83 & 1.94 & 13.79 & 7.54 & 1.99 & 1.81 & 2.72 & 1.65 & 22.41 & 5.08 & 3.24 & 6.19 \\ 
        \gemmanine & 1.84 & 1.20 & 8.39 & 4.24 & 2.61 & 0.73 & 2.99 & 1.21 & 6.58 & 3.96 & 1.85 & 3.31 \\ 
        \gemmats & 1.34 & 1.21 & 8.41 & \textbf{10.17} & 1.72 & 1.04 & 2.39  &  1.28 & 23.18 & 3.47  & 3.43 & 5.64 \\ 
        \llamas & 2.23 & 2.74 & 10.12 & 5.73 & 3.72 & 3.03 & 2.75 & 2.87 & 21.61 & 4.71 & 3.59 & 6.09 \\         
        \llamax & 5.29 & 4.90 & 18.11 & 10.16 & 3.38 & 2.54 & 3.06 & 0.00 & 31.25 & 7.46 & 4.40 & 8.74 \\ 
        \aya & \textbf{6.44} & \textbf{5.58} & \textbf{19.17} & 4.70 & 4.71 & 2.80 & \textbf{7.06} & \textbf{6.12} & 19.17 & \textbf{9.35} & \textbf{4.80} & 8.41 \\ 
        \gptf & 5.63 & 0.03 & 16.94 & 3.27 & \textbf{6.38} &  \textbf{4.70} & 6.00 & 3.88 & \textbf{50.39} & 8.73 & 2.97 & \textbf{10.80} \\ 
        \bottomrule
    \end{tabular}
    }
    \caption{ChrF Generation Scores. For \textit{native}, descriptions were provided in the native language, while for \textit{English}, descriptions were given in English to generate proverbs in each language. Native and English choice averages do not include the English language}
    \label{tab:generation-scores}
\end{table*}

\paragraph{Monolingual vs Cross-lingual understandings} 
We evaluate both monolingual and cross-lingual understanding by using native and English choices. The results indicate that in most cases the models demonstrate more robust performance in monolingual tasks than in cross-lingual ones, except for \gptf. Sensitivity to choice order is also less apparent when using monolingual (native) choices, as shown in Table \ref{tab:shuffle_stuff}.


\begin{table*}[!ht]
    \centering
    \resizebox{0.85\linewidth}{!}{%
    \begin{tabular}{l|ll|ll|ll|ll}
        \toprule
         \multirow{2}{4em}{\textbf{Model Name}} & \multicolumn{2}{|c|}{\textbf{Amharic} } & \multicolumn{2}{|c|}{\textbf{Afaan Oromo} } & \multicolumn{2}{|c|}{\textbf{Tigrinya} } & \multicolumn{2}{|c}{\textbf{Average}} \\ 
        \textbf{} & \textit{native} & \textit{english} & \textit{native} & \textit{english} & \textit{native} & \textit{english} & \textit{native} & \textit{english}   \\  \midrule
        \textbf{\llamathree} & ~ & ~ & ~ & ~ & ~ & ~ & ~ & ~   \\ 
        \textit{native proverb} & 24.72 & 24.98 & 32.54 & 25.37 & 26.93 & 29.83 & 28.06 & 26.73  \\ 
        \textit{translated proverb} & 32.10 & 27.33 & 26.49 & 32.37 & 23.51 & 32.37 & 27.37 & 30.69  \\ \hline
        \textbf{\gemmanine} & ~ & ~ & ~ & ~ & ~ & ~ & ~ & ~  \\ 
         \textit{native proverb} & 31.06 & 30.85 & 29.22 & 26.43 & 29.82 & 30.88 & 30.03 & 29.39  \\ 
         \textit{translated proverb} & 27.13 & 33.68 & 27.09 & 38.98 & 28.25 & 34.12 & 27.49 & 35.59  \\ \hline
        \textbf{\gemmats} & ~ & ~ & ~ & ~ & ~ & ~ & ~ & ~  \\ 
        \textit{native proverb} & 35.06 & 36.30 & 34.99 & 27.69 & 32.39 & 33.95 & 34.15 & 32.65 \\ 
        \textit{translated proverb} & 31.10 & 34.99 & 31.04 & 33.34 & 30.32 & 34.04 & 30.82 & 34.12  \\ \hline
        \textbf{\llamas} & ~ & ~ & ~ & ~ & ~ & ~ & ~ & ~ \\ 
        \textit{translated proverb} & 41.67 & 37.61 & 32.67 & 27.96 & 36.49 & 30.96 & 36.94 & 32.18  \\ 
        \textit{translated proverb} & 42.15 & 38.44 & 31.41 & 43.63 & 32.46 & 34.65 & 35.34 & 38.91  \\ \hline 
        \textbf{\llamax} & ~ & ~ & ~ & ~ & ~ & ~ & ~ & ~  \\ 
        \textit{native proverb} & 28.99 & 25.38 & 31.94 & 25.77 & 29.21 & 28.25 & 30.05 & 26.47  \\ 
        \textit{translated proverb} & 28.17 & 27.33 & 28.75 & 29.48 & 26.93 & 30.44 & 27.95 & 29.08  \\ \hline
        \textbf{\aya} & ~ & ~ & ~ & ~ & ~ & ~ & ~ & ~  \\ 
        \textit{native proverb} & 48.21 & 52.38 & 49.40 & 32.8 & 42.19 & 55.09 & 46.6 & 46.76  \\ 
        \textit{translated proverb} & 40.57 & 41.27 & 46.48 & 41.77 & 36.76 & 44.91 & 41.27 & 42.65  \\ \hline 
        \textbf{\gptf} & ~ & ~ & ~ & ~ & ~ & ~ & ~ & ~  \\ 
        \textit{native proverb} &40.19  &46.24  &49.01  &50.80  &32.37  &35.00  &40.52 &44.01  \\ 
        \textit{translated proverb} &59.40  &39.06  &42.57  &44.15  &49.34  &34.73  &50.43 & 39.31 \\ \hline
        \textbf{} & ~ & ~ & ~ & ~ & ~ & ~ & ~ & ~  \\  
        \textbf{Average native } & 34.95 & 31.02 & 32.27 & 26.64 & 30.97 & 30.77 & 32.73 & 29.48  \\ 
        \textbf{Average translated } & 33.54 & 33.84 & 31.88 & 36.6 & 29.71 & 35.09 & 31.71 & 35.18  \\ 
        \bottomrule
    \end{tabular}
    }
    \caption{Accuracy scores of proverb translate-test. Can translating proverbs using NLLB-200 3.3B \cite{nllb2022} improve the performance of task 1 (\textit{ meaning multiple choice task})?  This experiment covers languages supported by NLLB.}
    \label{table:translate-test}
\end{table*}

\paragraph{Does translating proverbs into English improve low-resource language performance?}
Table \ref{table:translate-test} shows the effect of translating proverbs written in low-resource languages into English. As we can see from the average results, translating proverbs into English does not significantly help models. 


\subsection{Task 2: Proverb Fill in the Blank}

\paragraph{Zero-shot \& Few-shot results of fill the blank}

Looking at Figure \ref{fig:zero-shot-all}, we observe that all models perform poorly in the fill-in-the-blank task, with the exception of Ge’ez and English for \gptf. The task appears to be easily solvable in English, probably because of the strong focus on English in these models. The examples presented demonstrate a modest performance improvement for open-source models, whereas \gptf shows less benefits from few-shot examples.



\subsection{Task 3: Proverb Generation Task}
\paragraph{Can LLMs generate coherent proverbs for a given description in low-resource language?} 
Table \ref{tab:generation-scores} shows the ability of the models to generate proverbs for a given description in \textbf{\textit{native}} language and in \textbf{\textit{English}}. \texttt{LlaMA} models show strong generation ability when the description is given in the native language, and \gemmats becomes competitive when the description is given in the English language, looking at the average scores. There is a huge difference between languages that use Latin script and others that use Ge'ez script.

\textbf{English vs. Native Descriptions} In most cases, models are more likely to generate proverbs in native languages when provided with native descriptions compared to English descriptions. This is because, when given native input, the models tend to anchor their generation around key culturally specific terms or phrases. This context-sensitive approach often results in more accurate and culturally relevant proverb generation, as the models are better able to capture nuances inherent in the native language.



\subsection{General Takeaways}
\paragraph{Building Models Optimized for Multilingual Functionality} Designing an effective tokenizer is crucial, as it serves as a strong foundation for developing more advanced LLMs.
\paragraph{Size Is Not Always the Answer} Models with better tokenizers and fine-tuned on carefully curated datasets can be competitive with larger models.
\paragraph{Monolingual vs. Cross-Lingual Evaluations} When designing LLM evaluations, it is crucial to consider the differences between monolingual and cross-lingual properties.
\paragraph{Subject vs. Language Understanding} Distinguishing between language and subject understanding is crucial in LLM evaluation.
\paragraph{Translate Test Experiment} Creating a benchmark that captures the cultural and linguistic nuances of a language is crucial for evaluating LLMs. This ensures that language understanding assessments are robust and not artificially inflated by simple translation systems.
\paragraph{Distinct Patterns in Ge’ez Proverbs} We carefully analyzed the linguistic patterns in Ge’ez and observed distinct behaviors in certain tests. Our findings suggest the following characteristics: (1) Ge’ez proverbs are predominantly derived from biblical sources, making them more predictable. (2) Instead of focusing on everyday activities, they emphasize spiritual traditions and customs, providing limited contextual diversity. (3) Ge’ez proverbs are generally shorter and more predictable than those in other languages. (4) The dataset used for Ge’ez proverbs was sourced from a single collection, increasing the likelihood of its inclusion in common LLM training datasets.

\section{Conclusion}
\label{Conclusion and Future Work}


In this work, we explore the challenges of LLM evaluation for low-resource language understanding. We also introduce a \proverbeval, LLM evaluation benchmark for low-resource language based on proverbs to focus on low-resource language understanding in culture-specific scenarios. Our results indicate that LLMs still significantly underperform in non-English languages when it comes to understanding proverbs, as compared to their performance in English. We observed that prompting LLMs in their native languages leads to lower accuracy, and the models have high sensitivity to the order in which choices are presented. In the fill-in-the-blank task, few-shot prompting showed minimal improvement. In generative tasks, LLMs perform better when descriptions are provided in native languages. 

In benchmarks focused on cultural understanding, some results may not be transferable to other languages or to broader evaluations. However, this highlights the need for specialized evaluations that capture cultural nuances to ensure that LLMs demonstrate true language understanding.

\section*{Acknowledgment}

We thank Hellina Hailu Nigatu for her feedback and input on earlier versions of this work.

\section*{Limitations}

\paragraph{Should open-source and closed-source model results be reported together?} Open-source models can be easily evaluated using log-likelihood scores. However, this approach is not feasible for closed-source models. As a result, we converted all tasks to generation-based evaluation for closed-source models, a method widely adopted across various evaluation benchmarks. Despite its popularity, these results should not be considered directly comparable. The performance of closed-source models highly depends on the specific verbalizer (tool used to extract answers from long generations) used for each task.

\paragraph{Label-based vs sequence-based evaluations} An interesting question explored by \citet{lyu-etal-2024-beyond} is whether to evaluate large language models (LLMs) based on the probability assigned to the multiple-choice letter (e.g., "A" for the first option) or the content of the choice itself. In this work, given the extensive number of experiments we conducted, we opted to use label-based evaluation.

\paragraph{Language coverage} The scope of this work includes a limited number of languages, primarily constrained by the availability of volunteer native speakers and resource limitations. Expanding the language coverage to include a broader range of cultures and languages would significantly enhance the utility of the benchmark, making it a more comprehensive tool for evaluating model performance across diverse linguistic and cultural contexts.

\paragraph{Limited LLMs evaluation} The number of LLMs evaluated in this work is limited. Expanding the study to include a broader range of both open-source and closed-source models could provide deeper insights. Additionally, an important avenue for future research is exploring how this type of language understanding can inform the development of more robust multilingual models. However, this particular question falls outside the scope of the present study.
 


\bibliography{custom}

\appendix
\section{Details of covered languages }
\label{app:lang}
There are more than 2000 languages spoken in the African continent, and more than 80 of them are spoken in Ethiopia\footnote{https://www.statista.com/statistics/1280625/number-of-living-languages-in-africa-by-country/}. Amharic, Afaan Oromo, and Tigrinya are the top languages in Ethiopia by the number of speakers. Ge'ez language is also known as Ethiopic script, the origin of Amharic and Tigrinya languages.
~\\\textbf{Amharic} (amh): is a Semitic language written in Ge'ez script, which consists of 33 primary characters, each with seven vowel sequences. It is the second most widely spoken Semitic language, next to Arabic.
~\\\textbf{Afaan Oromo} (orm): is an Afro-Asiatic language written in Latin script. It is the most widely spoken language in Ethiopia and the third most widely spoken in Africa, next to the Arabic and Hausa languages
~\\\textbf{Tigrinya} (tir): is a Semitic language spoken in the Northern part of Ethiopia and Eritrea. The language uses Ge'ez script with additional Tigrinya alphabets and it is the fourth widely spoken language in Ethiopia next to Somali \cite{Eberhard2023}.
~\\\textbf{Ge'ez} (gez): is a language of Ethiopia that is used only as a second language and does not have an ethnic community. It belongs to the Afro-Asiatic language family.
~\\\textbf{English} (eng): we have created a new proverb dataset for the English language. The proverb descriptions of the other languages also have English descriptions for parallel evaluation.
\onecolumn

\section{Prompts for zero-shot and ICL}
\label{app:prompt-details}
Table \ref{tab:amh} presents all prompts used in the three proposed tasks: Task1: multiple choice, Task2: fill in the blank, and Task: proverb description generation, respectively.
\begin{table}[ht!]
\scalebox{0.7}{
\begin{tabular}{p{210mm}}
\toprule
\textit{English multiple choice prompts} \\ 
\textbf{Prompt 1}: \texttt{You are LLM capable of understanding \{language\} language. I will give you a prompt and a list of descriptions that have the same meaning. Return a letter for the correct choice among four choices given} \\ 
\\ 
\textbf{Prompt 2}: \texttt{You are LLM capable of understanding language. I will give you a prompt and a list of descriptions that have the same meaning. Return a letter for the correct choice among four choices given} \\ 
\\
\textbf{Prompt 3}: \texttt{ Which choice are similar?} \\ 
\midrule

\textit{Afaan Oromo multiple choice prompts} \\ 
\textbf{Prompt 1}: \texttt{ Ati LLM dandeetti Afan \{language\} hubachuu qabdudha. Gaaffii fi fillannoowwan hiikaa/eergaa ibsan sif nan laadha.         Filannoowwan afur keennaman keessaa quubee deebiin sirri irra jiru naaf deebisii.} \\ 
\\ 
\textbf{Prompt 2}: \texttt{Ati LLM dandeetti Afan hubachuu qabdudha. You are LLM capable of understanding language. Gaaffii fi fillannoowwan hiikaa/eergaa ibsan sif nan laadha. Filannoowwan afur keennaman keessaa quubee deebiin sirri irra jiru naaf deebisii.} \\ 
\\
\textbf{Prompt 3}: \texttt{Fiilannoowwan armaan gadii keessaa kamtuu hiika/eergaa walfakkaataa qaba} \\ 
\midrule

\textit{Amharic multiple choice prompts} \\ 
\textbf{Prompt 1}: \selectlanguage{ethiop} 'aneta \selectlanguage{english} \{language\} \selectlanguage{ethiop} quAnequA maradAte yamete^cele yakomepitare mA^sene nahe:: bamaqa.tle mesAlEyAwi 'anagAgare 'enA teregumo^ce sa.tehAlahu katasa.tute 'arAte mere^CAwo^ce makAkale latekekela~nAwe mere^CA fidale yemalesu\selectlanguage{english}\\ 
\\                
\textbf{Prompt 2}: \selectlanguage{ethiop} 'aneta quAnequA maradAte yamete^cele yakomepitare mA^sene nahe:: bamaqa.tle mesAlEyAwi 'anagAgare 'enA teregumo^ce sa.tehAlahu ketasa.tute 'arAte mere^CAwo^ce makAkale latekekela~nAwe mere^CA fidale yemalesu::\selectlanguage{english}  \\ 
 \\                
\textbf{Prompt 3}: \selectlanguage{ethiop}yate~nAwewe mere^CA tamasAsAye nawe?\selectlanguage{english}\\ 
\midrule

\textit{Tigrinya multiple choice prompts} \\ 
\textbf{Prompt 1}: \selectlanguage{ethiop} nese_kA \selectlanguage{english} \{language\} \selectlanguage{ethiop} zetabehAla  quAnequA meredA'e 'A'qemi zalakA Abeyi nAye quAnequA sele.tune 'I_kA:: zerezere tamasAsAli teregume zalawome magela.sitAte mese ma.taya'qetA kehebakA 'eya:: kAbetome 'ArebA'eta mere^CAtAte 'qenue ze_kona malesi ze.hAza mamAra.ci fidale mera.ce::
\selectlanguage{english} \\ 
\\                
\textbf{Prompt 2}: \selectlanguage{ethiop} nese_kA quAnequA meredA'e 'A'qemi zalakA Abeyi nAye quAnequA sele.tune 'I_kA:: zerezere tamasAsAli teregume zalawome magela.sitAte mese ma.taya'qetA kehebakA 'eya:: kAbetome 'ArebA'eta mere^CAtAte 'qenue ze_kona malesi ze.hAza mamAra.ci fidale mera.ce:: \selectlanguage{english}\\ 
 \\ 
\textbf{Prompt 3}: \selectlanguage{ethiop}'AyanAye mere^CA 'eyu tamasAsAli?\selectlanguage{english}\\ 
\midrule

\textit{Ge'ez multiple choice prompts} \\ 
\textbf{Prompt 1}: \selectlanguage{ethiop} 'aneta weetu kine wazata'amere lesAna\selectlanguage{english} \{language\}:: \selectlanguage{ethiop} 
'emezatA.hetu zatadAlawe 'emArEyAte ba_hreyo 'awesee (_hraye) fidala zasene'ewe 'emezatadalawa 'anegAra mesAlE::\selectlanguage{english} \\ 
\\                
\textbf{Prompt 2}: \selectlanguage{ethiop} 'aneta weetu kine wazata'amere lesAne:: 'emezatA.hetu zatadAlawe 'emArEyAte ba_hreyo 'awesee (_hraye) fidala zasene'ewe 'emezatadalawa 'anegAra mesAlE::\selectlanguage{english}\\ 
 \\      
\textbf{Prompt 3}: \selectlanguage{ethiop}
'aye we'etu 'anegAra mesAlE eruya zakona 'awe zayeqarebe mesela 'elatadalawe 'anegAra mesAlEyAte 'emezatA.hetu?\selectlanguage{english}\\ 
\midrule
\textit{Fill the blank prompt} \\ 
\textbf{English}: \texttt{ You are LLM capable of understanding \{language\} language. Given a proverb, can you fill the blank with an appropriate word from the choices? blank is shown with '\_\_\_'. }\\
\texttt{\{language\} Proverb: \{Proverb\}} \\                
\texttt{Choices:}  \\ 
\texttt{      A: \{A\}}  \\ 
\texttt{      B: \{B\}}  \\ 
\texttt{      C: \{C\}}  \\ 
\texttt{      D: \{D\}}  \\ 
\texttt{} \\ 
\texttt{Answer:}
\\ 
\midrule
\textit{Proverb generation prompt} \\ 
\texttt{ You are LLM capable of understanding \{language\} language. Based on the detailed description provided in \{source\_language\}, generate an appropriate proverb in \{target\_language\} that captures the essence and meaning of the context.}\\
\\
\texttt{\{source\_language\} Description: \{Description\}}\\
\texttt{\{target\_language\} Proverb: }
\\ 

\bottomrule
\end{tabular}
}
\caption{Prompts (in five languages) used for decoder-only zero-shot and in-context learning experiments} 
\label{tab:amh}
\end{table}
\FloatBarrier

\FloatBarrier

\FloatBarrier
\FloatBarrier

\section{Multiple choice detail results }
\label{app:mc}

In Table \ref{tab:choice_order}, we present an alternative analysis of choice order variance. We averaged the results from the first three randomly shuffled prompts and compared them against instances where the correct choice appeared as either the first or last option. The numbers reflect the deviation from the average random shuffle baseline. For example, a value of +33.92 indicates an increase in accuracy by 33.92 percentage points, while -19.63 signifies a decrease of 19.63 points relative to the baseline.

\begin{table*}[ht!]
    \centering
    \resizebox{\linewidth}{!}{%
    \begin{tabular}{l|ll|ll|ll|ll|l|lll}
        \toprule
        \multirow{2}{8em}{\textbf{Model Name} \\ \textit{shuffling strategy}} & \multicolumn{2}{|c|}{\textbf{Amharic} } & \multicolumn{2}{|c|}{\textbf{Afaan Oromo} } & \multicolumn{2}{|c|}{\textbf{Tigrinya} } & \multicolumn{2}{|c|}{\textbf{Ge'ez} } & \textbf{English} & \multicolumn{3}{|c}{\textbf{average}} \\ 
        \textbf{} & \textit{native} & \textit{english}  & \textit{native} & \textit{english} & \textit{native} & \textit{english}  & \textit{native} & \textit{english}  &  & \textit{native} & \textit{english}  & \textit{all} \\ \midrule
        \textbf{\llamathree} & ~ & ~ & ~ & ~ & ~ & ~ & ~ & ~ & ~ & ~ & ~ & ~ \\ 
         \textit{3 random shuffle Avg.}& 26.86 & 26.98 & 31.54 & 25.77 & 29.3 & 26.05 & 30.34 & 27.67 & 50.73 & 29.51 & 26.61 & 30.58 \\ 
        \textit{All answers A Diff} & +32.02 & +46.93 & +38.38 & +54.31 & +26.49 & +54.21 & +38.78 & +49.96 & +38.74 & +33.92 & +51.35 & +42.20 \\ 
        \textit{All answers D Diff} & -19.63 & -17.04 & -14.81 & -20.79 & -25.35 & -23.16 & -25.96 & -23.2 & -27.16 & -21.44 & -21.05 & -21.9 \\ \hline
        \textbf{\gemmanine} & ~ & ~ & ~ & ~ & ~ & ~ & ~ & ~ & ~ & ~ & ~ & ~ \\ 
         \textit{3 random shuffle Avg.}& 29.68 & 33.89 & 31.54 & 28.89 & 29.47 & 27.46 & 40.4 & 27.37 & 64.23 & 32.77 & 29.4025 & 34.77 \\ 
        \textit{All answers A Diff} & +33.34 & +47.27 & +37.58 & +59.76 & +21.06 & +60.17 & +32.87 & +58.68 & +26.62 & +31.21 & +51.35 & +42.20 \\ 
        \textit{All answers D Diff} & -7.57 & -20.23 & -7.44 & -24.51 & -5.00 & -22.46 & -9.06 & -22.32 & -13.43 & -7.27 & -22.38 & -14.67 \\ \hline
        \textbf{\gemmats} & ~ & ~ & ~ & ~ & ~ & ~ & ~ & ~ & ~ & ~ & ~ & ~ \\ 
         \textit{3 random shuffle Avg.}& 34.64 & 34.57 & 36.25 & 29.02 & 28.16 & 29.91 & 40.02 & 29.82 & 66.13 & 34.7675 & 30.83 & 36.50\\ 
        \textit{All answers A Diff} & +30.65 & +35.00 & +26.1 & +45.48 & +26.84 & +43.77 & +32.79 & +45.18 & +24.26 & +29.09 & +42.35 & +34.45 \\ 
        \textit{All answers D Diff} & -15.43 &	-13.87	&-10.55	&-19.26 &	-10.00	&-22.8 &	-12.83	&-21.14	&-11.9& -12.20 & -19.27 & -15.30 \\ \hline
        \textbf{\llamas} & ~ & ~ & ~ & ~ & ~ & ~ & ~ & ~ & ~ & ~ & ~ & ~ \\ 
         \textit{3 random shuffle Avg.}& 41.94 & 40.3 & 35.13 & 31.51 & 38.16 & 30.18 & 61.44 & 30.26 & 74.52 & 44.17 & 33.06 & 42.60 \\ 
        \textit{All answers A Diff} & +8.68 & +12.91 & +21.64 & +18.89 & +3.68 & +28.5 & +10.22 & +28.42 & +5.34 & +11.06 & +22.18 & +15.36 \\ 
        \textit{All answers D Diff} & -13.43 & -18.97 & -13.42 & -16.77 & -17.63 & -20.71 & -18.81 & -20.00 & -2.67 & -15.82 & -19.11 & -15.82 \\ \hline
        \textbf{\llamax} & ~ & ~ & ~ & ~ & ~ & ~ & ~ & ~ & ~ & ~ & ~ & ~ \\ 
         \textit{3 random shuffle Avg.}& 33.95 & 25.33 & 31.48 & 26.29 & 30.79 & 27.63 & 33.18 & 27.19 & 39.51 & 32.35 & 26.61 & 30.55 \\ 
        \textit{All answers A Diff} & +3.03 & +39.06 & +16.33 & +45.82 & -5.00 & +51.84 & +6.68 & +53.34 & +36.46 & +5.26 & +47.51 & +27.50 \\ 
        \textit{All answers D Diff} & +0.55 & -14.77 & -9.57 & -21.91 & +6.84 & -23.16 & -2.53 & -22.98 & -15.48 & -1.17 & -20.71 & -11.44 \\ \hline
        \textbf{\aya} & ~ & ~ & ~ & ~ & ~ & ~ & ~ & ~ & ~ & ~ & ~ & ~ \\ 
         \textit{3 random shuffle Avg.}& 51.24 & 54.98 & 51.06 & 32.8 & 43.16 & 55.35 & 78.88 & 55.88 & 80.78 & 56.09 & 49.75 & 56.01 \\ 
        \textit{  All answers A Diff} & +10.74 & +12.93 & +3.52 & +11.82 & +8.16 & +12.28 & +6.83 & +14.38 & +1.56 & +7.31 & +12.85& +9.13 \\ 
        \textit{All answers D Diff} & +5.99 & +1.75 & +0.93 & -1.53 & +7.37 & -4.82 & +2.92 & -6.67 & +0.00 & +4.30 & -2.82 & +0.66 \\ 
         \hline 
        \textbf{\gptf} & ~ & ~ & ~ & ~ & ~ & ~ & ~ & ~ & ~ & ~ & ~ & ~ \\ 
         \textit{3 random shuffle Avg.}& 54.13 & 66.39 & 76.73 & 80.35 & 44.30 & 42.28 & 87.48 & 76.62 & 99.46 &	65.66 &	69.41 &
         69.75 \\ 
        \textit{  All answers A Diff} & -1.65 & -8.21 & +3.95 & -1.27 & +1.75 & +6.67 & +5.84 & -23.99 & +0.08 & +2.47 & -6.70 &	-1.87 \\ 
        \textit{All answers D Diff} & -5.99 & -8.21 & -3.62 & -4.25 & -7.19 & -2.28 & -8.45 & -36.88 & -0.38 & -6.31 & -12.90 & -8.58 \\ 
        \bottomrule
    \end{tabular}
    }
    
    \caption{Zero-shot scores of task 1 (\textit{ meaning multiple choice task}) across all models for \textbf{\textit{native}} choices and \textbf{\textit{english}} choices.  The first row shows the average across \textit{\textbf{three different random shuffles}} of the choice order. We compare this with providing the \textbf{\textit{correct choice at choice "A" (first choice)}} or providing the \textbf{\textit{correct choice at choice "D" (last choice)}}. For choices "A" and "D," the closer the numbers are to zero, the better since we don't see huge variance from the shuffle. }
    \label{tab:choice_order}
\end{table*}

\clearpage
\section{Choice sensitivity }
\label{app:choice-sensetivity}

In Table \ref{tab:choicesheffle}, we present three different results based on the order of choices. To provide detailed insights, we have listed all outcomes, revealing that random shuffling yields consistent results. However, when the correct answer is consistently positioned as either the first option (‘A’) or the last option (‘D’), we observe significant variations in performance.

\begin{table*}[!ht]
    \centering
    \resizebox{\linewidth}{!}{%
    \begin{tabular}{l|ll|ll|ll|ll|l|l|l}
    \toprule
        \multirow{2}{4em}{} & \multicolumn{2}{|c|}{\textbf{Amharic}}  & \multicolumn{2}{|c|}{\textbf{Afaan Oromoo}}& \multicolumn{2}{|c|}{\textbf{Tigrinya}} & \multicolumn{2}{|c|}{\textbf{Ge'ez}} & \textbf{English} & \multirow{2}{6em}{\textbf{Model Name}} & \multirow{2}{6em}{\textbf{Choice Order}} \\ 
         &\textit{native} & \textit{english} & \textit{native} & \textit{english} & \textit{native} & \textit{english} & \textit{native} & \textit{english} & \textit{native} &  & ~ \\ \midrule
        \textbf{shuffle 1} & 27.48 & 24.22 & 32.07 & 25.90 & 27.11 & 28.95 & 32.26 & 29.74 & 51.03 &  & random \\ 
        \textbf{shuffle 2} & 25.21 & 27.54 & 31.47 & 26.10 & 31.05 & 22.63 & 29.26 & 23.68 & 49.89 &  & random \\ 
        \textbf{shuffle 3} & 27.89 & 29.19 & 31.08 & 25.3 & 29.74 & 26.58 & 29.49 & 29.58 & 51.26 & \llamathree & random \\ 
        \textbf{shuffle 4} & 58.88 & 73.91 & 69.92 & 80.08 & 55.79 & 80.26 & 69.12 & 77.63 & 89.47 &  & A \\ 
        \textbf{shuffle 5} & 7.23 & 9.94 & 16.73 & 4.98 & 3.95 & 2.89 & 4.38 & 4.47 & 23.57 &  & D \\ 
        \textbf{} & ~ & ~ & ~ & ~ & ~ & ~ & ~ & ~ & ~ & ~ & ~ \\ 
        \textbf{shuffle 1} & 30.79 & 31.47 & 30.88 & 28.09 & 29.21 & 31.58 & 37.1 & 31.58 & 64.53 &  & random \\ 
        \textbf{shuffle 2} & 27.27 & 34.58 & 33.07 & 30.88 & 28.16 & 24.21 & 42.17 & 23.68 & 64.53 &  & random \\ 
        \textbf{shuffle 3} & 30.99 & 35.61 & 30.68 & 27.69 & 31.05 & 26.58 & 41.94 & 26.84 & 63.62 & \gemmanine & random \\ 
        \textbf{shuffle 4} & 63.02 & 81.16 & 69.12 & 88.65 & 50.53 & 87.63 & 73.27 & 86.05 & 90.85 &  & A \\ 
        \textbf{shuffle 5} & 22.11 & 13.66 & 24.10 & 4.38 & 24.47 & 5 & 31.34 & 5.05 & 50.80 &  & D \\ 
        \textbf{} & ~ & ~ & ~ & ~ & ~ & ~ & ~ & ~ & ~ & ~ & ~ \\ 
        \textbf{shuffle 1} & 34.30 & 35.40 & 35.25 & 29.68 & 28.42 & 35.79 & 41.71 & 34.74 & 65.90 & & random \\ 
        \textbf{shuffle 2} & 31.61 & 34.99 & 37.65 & 28.88 & 26.58 & 26.05 & 36.64 & 26.84 & 67.96 & & random \\ 
        \textbf{shuffle 3} & 38.02 & 33.33 & 35.86 & 28.49 & 29.47 & 27.89 & 41.71 & 27.89 & 64.53 & \gemmats& random \\ 
        \textbf{shuffle 4} & 65.29 & 69.57 & 62.35 & 74.5 & 55.00 & 73.68 & 72.81 & 75.00 & 90.39 & & A \\ 
        \textbf{shuffle 5} & 19.21 & 20.70 & 25.70 & 9.76 & 18.16 & 7.11 & 27.19 & 8.68 & 54.23 & & D \\ 
        \textbf{} & ~ & ~ & ~ & ~ & ~ & ~ & ~ & ~ & ~ & ~ & ~ \\ 
        \textbf{shuffle 1} & 41.53 & 37.47 & 33.86 & 31.27 & 39.47 & 30.26 & 61.06 & 28.95 & 75.29 &  & random \\
        \textbf{shuffle 2} & 41.94 & 40.17 & 36.06 & 30.68 & 37.63 & 30.53 & 60.14 & 30.26 & 75.97 &  & random \\ 
        \textbf{shuffle 3} & 42.36 & 43.27 & 35.46 & 32.58 & 37.37 & 29.74 & 63.13 & 31.58 & 72.31 & \llamas & random \\ 
        \textbf{shuffle 4} & 50.62 & 53.21 & 56.77 & 50.4 & 41.84 & 58.68 & 71.66 & 58.68 & 79.86 &  & A \\ 
        \textbf{shuffle 5} & 28.51 & 21.33 & 21.71 & 14.74 & 20.53 & 9.47 & 42.63 & 10.26 & 71.85 &  & D \\
        \textbf{} & ~ & ~ & ~ & ~ & ~ & ~ & ~ & ~ & ~ & ~ & ~ \\ 
        \textbf{shuffle 1} & 33.88 & 23.81 & 30.88 & 26.69 & 30.79 & 31.05 & 30.18 & 29.74 & 38.67 &  & random \\ 
        \textbf{shuffle 2} & 33.68 & 27.12 & 30.88 & 27.09 & 32.37 & 23.42 & 36.18 & 24.21 & 43.25 &  & random \\ 
        \textbf{shuffle 3} & 34.30 & 25.05 & 32.67 & 25.10 & 29.21 & 28.42 & 33.18 & 27.63 & 36.61 & \llamax & random \\ 
        \textbf{shuffle 4} & 36.98 & 64.39 & 47.81 & 72.11 & 25.79 & 79.47 & 39.86 & 80.53 & 75.97 &  & A \\ 
        \textbf{shuffle 5} & 34.50 & 10.56 & 21.91 & 4.38 & 37.63 & 4.47 & 30.65 & 4.21 & 24.03 &  & D \\ 
        \textbf{} & ~ & ~ & ~ & ~ & ~ & ~ & ~ & ~ & ~ & ~ & ~ \\ 
        \textbf{shuffle 1} & 50.00 & 55.62 & 50.20 & 33.47 & 42.89 & 57.89 & 78.57 & 57.89 & 80.09 &  & random \\ 
        \textbf{shuffle 2} & 51.45 & 54.66 & 51.20 & 33.47 & 43.42 & 53.95 & 78.34 & 53.95 & 80.32 &  & random \\ 
        \textbf{shuffle 3} & 52.27 & 54.66 & 51.79 & 31.47 & 43.16 & 54.21 & 79.72 & 55.79 & 81.92 & \aya & random \\ 
        \textbf{shuffle 4} & 61.98 & 67.91 & 54.58 & 44.62 & 51.32 & 67.63 & 85.71 & 70.26 & 82.38 &  & A \\ 
        \textbf{shuffle 5} & 57.23 & 56.73 & 51.99 & 31.27 & 50.53 & 50.53 & 81.8 & 49.21 & 80.78 &  & D \\ 
        \textbf{} & ~ & ~ & ~ & ~ & ~ & ~ & ~ & ~ & ~ & ~ & ~ \\ 
        \textbf{shuffle 1} &53.51 &67.29 &76.2 &79.88 &45.53 &53.16 &87.33 &51.05 & 99.54 &  & random \\ 
        \textbf{shuffle 2} &53.31 &65.01 &76.69 &80.28 &42.89 &47.11 &87.10 & 90.09 &99.31 &  & random \\ 
        \textbf{shuffle 3} &55.58 &66.87 &77.29 &80.88 &44.47 &52.89 &88.02 &88.71 &99.54 & \gptf & random \\ 
        \textbf{shuffle 4}&52.48 &58.18 &80.68 &79.08 &46.05 &48.95 &93.32 &52.63 &99.54 &  & A \\ 
        \textbf{shuffle 5} &48.14 &58.18 & 73.11 &76.10 &37.11 &40.00 &79.03 &39.74 &99.08 &  & D \\ 
        \bottomrule
        
    \end{tabular}
    
    }
    \caption{Accuracy scores for \textit{Task 1: Meaning Multiple Choice } task given three different randomly shuffled choices and when we make the choices first or last choice for the whole dataset.}
    \label{tab:choicesheffle}
\end{table*}

\clearpage
\section{English Prompt Sensitivity for Multiple choice}
\label{app:Prompt Sensitivity}
Table Table \ref{tab:englishprompts} provides a detailed analysis of the model’s sensitivity and accuracy when responding to three distinct prompts in English. The table is focused on assessing how well the model adapts to varying prompt formulations and maintains accuracy in its responses. Model outputs can be different depending on the prompt used in our task and on Table \ref{tab:englishprompts} and Table \ref{tab:nativeprompts}. We explored both native and English prompts.

\begin{table*}[!ht]
    \centering
    
    \resizebox{\linewidth}{!}{%
    \begin{tabular}{l|ll|ll|ll|ll|l|l}
    \toprule
    \multirow{2}{4em}{} & \multicolumn{2}{|c|}{\textbf{Amharic}}  & \multicolumn{2}{|c|}{\textbf{Afaan Oromoo}}& \multicolumn{2}{|c|}{\textbf{Tigrinya}} & \multicolumn{2}{|c|}{\textbf{Ge'ez}} & \textbf{English} & \multirow{2}{6em}{\textbf{Model Name}}  \\ 
         &\textit{native} & \textit{english} & \textit{native} & \textit{english} & \textit{native} & \textit{english} & \textit{native} & \textit{english} & \textit{native} &  \\ \midrule
        \textbf{Prompt 1} & 26.65 & 24.22 & 31.67 & 25.9 & 27.89 & 28.95 & 32.26 & 29.74 & 51.26 &  \\ 
        \textbf{Prompt 2} & 23.55 & 26.29 & 33.27 & 24.7 & 25.26 & 29.74 & 29.49 & 29.95 & 52.63 & \llamathree \\ 
        \textbf{Prompt 3} & 23.97 & 24.43 & 32.67 & 25.5 & 27.63 & 30.79 & 28.57 & 28.11 & 44.39 &  \\ 
        \textbf{} & ~ & ~ & ~ & ~ & ~ & ~ & ~ & ~ & ~ & ~ \\ 
        \textbf{Prompt 1} & 30.79 & 31.47 & 30.88 & 28.09 & 29.21 & 31.58 & 37.1 & 44.01 & 64.53 &  \\ 
        \textbf{Prompt 2} & 30.79 & 28.78 & 26.89 & 26.29 & 29.47 & 30.26 & 36.41 & 44.93 & 59.04 & \gemmanine \\ 
        \textbf{Prompt 3} & 31.61 & 32.3 & 29.88 & 24.9 & 30.79 & 30.79 & 40.78 & 48.85 & 66.36 &  \\ 
        \textbf{} & ~ & ~ & ~ & ~ & ~ & ~ & ~ & ~ & ~ & ~ \\ 
        \textbf{Prompt 1} & 34.3 & 35.4 & 38.25 & 29.68 & 28.42 & 35.79 & 41.71 & 38.49 & 65.9 &  \\ 
        \textbf{Prompt 2} & 35.33 & 35.4 & 29.47 & 26.29 & 35.06 & 32.89 & 40.09 & 44.01 & 64.99 & \gemmats \\ 
        \textbf{Prompt 3} & 35.54 & 38.1 & 37.25 & 27.09 & 33.68 & 33.16 & 43.78 & 45.62 & 73.46 &  \\ 
        \textbf{} & ~ & ~ & ~ & ~ & ~ & ~ & ~ & ~ & ~ & ~ \\ 
        \textbf{Prompt 1} & 41.74 & 37.47 & 33.47 & 31.08 & 37.11 & 31.05 & 61.06 & 49.77 & 75.06 &  \\ 
        \textbf{Prompt 2} & 42.36 & 40.17 & 32.87 & 27.69 & 36.58 & 30.79 & 57.14 & 48.39 & 75.06 & \llamas \\ 
        \textbf{Prompt 3} & 40.91 & 35.2 & 31.67 & 25.1 & 35.79 & 31.05 & 47.00 & 44.70 & 64.99 &  \\ 
        \textbf{} & ~ & ~ & ~ & ~ & ~ & ~ & ~ & ~ & ~ & ~ \\ 
        \textbf{Prompt 1} & 29.13 & 24.02 & 31.27 & 26.1 & 28.42 & 28.95 & 34.33 & 31.34 & 39.36 &  \\ 
        \textbf{Prompt 2} & 27.27 & 26.64 & 32.67 & 25.7 & 27.63 & 28.95 & 33.87 & 30.65 & 44.16 & \llamax \\ 
        \textbf{Prompt 3} & 30.58 & 25.47 & 31.87 & 25.5 & 31.58 & 26.84 & 36.87 & 30.18 & 44.62 &  \\ 
        \textbf{} & ~ & ~ & ~ & ~ & ~ & ~ & ~ & ~ & ~ & ~ \\ 
        \textbf{Prompt 1} & 50 & 55.69 & 50.2 & 33.47 & 42.89 & 57.89 & 78.57 & 83.64 & 80.09 &  \\ 
        \textbf{Prompt 2} & 48.35 & 54.24 & 50.6 & 32.27 & 41.84 & 56.58 & 74.65 & 83.87 & 79.41 & \aya \\ 
        \textbf{Prompt 3} & 46.28 & 47.2 & 47.41 & 32.67 & 41.84 & 50.79 & 72.81 & 79.95 & 72.77 &  \\  
        \textbf{} & ~ & ~ & ~ & ~ & ~ & ~ & ~ & ~ & ~ & ~ \\ 
        \textbf{Prompt 1} &62.81 &67.08 &78.29 &79.88 &46.58 &52.89 &31.57 &88.25 &99.50 &  \\ 
        \textbf{Prompt 2} &17.58 &70.39 &19.72 &72.31 &18.16 &51.84&16.82 &88.94 &99.50 &  \gptf \\ 
        \textbf{Prompt 3}  &0.00 &1.24& 0.00& 0.20&0.00 & 0.26&0.00 &0.69&70.71&  \\ 
        \bottomrule
    \end{tabular}
    }
    \caption{English Prompt sensitivity Accuracy results for three distinct prompts.}
    \label{tab:englishprompts}
\end{table*}

\clearpage
\section{Native Prompt Sensitivity for Multiple choice}
\label{tab:nativeprompts}
Table \ref{tab:nativeprompts} provides a detailed breakdown of results from three native (in-language) prompts used in a multiple-choice task. The prompts are designed in the respective native languages of the evaluation, and the task aims to assess the model’s performance in understanding and responding correctly to multiple-choice questions. 
\begin{table*}[!ht]
    \centering
    
    \resizebox{0.8\linewidth}{!}{%
    \begin{tabular}{l|ll|ll|ll|ll|l}
        \toprule
        \multirow{2}{4em}{} & \multicolumn{2}{|c|}{\textbf{Amharic}}  & \multicolumn{2}{|c|}{\textbf{Afaan Oromoo}}& \multicolumn{2}{|c|}{\textbf{Tigrinya}} & \multicolumn{2}{|c|}{\textbf{Ge'ez}} & \multirow{2}{6em}{\textbf{Model Name}} \\ 
        
        &\textit{native} & \textit{english} & \textit{native} & \textit{english} & \textit{native} & \textit{english} & \textit{native} & \textit{english} &  \\ \midrule
        \textbf{Prompt 1} & 33.06 & 26.71 & 27.09 & 25.3 & 27.37 & 26.84 & 26.96 & 24.19 &  \\ 
        \textbf{Prompt 2} & 34.09 & 26.29 & 24.9 & 24.7 & 27.11 & 23.68 & 26.73 & 25.12 & \llamathree \\ 
        \textbf{Prompt 3} & 27.48 & 26.29 & 26.69 & 24.9 & 26.84 & 24.74 & 25.58 & 21.66 &  \\ 
        \textbf{} & ~ & ~ & ~ & ~ & ~ & ~ & ~ & ~ & ~ \\ 
        \textbf{Prompt 1} & 34.09 & 38.72 & 25.3 & 27.09 & 28.95 & 27.11 & 26.5 & 26.04 &  \\ 
        \textbf{Prompt 2} & 28.31 & 36.85 & 24.9 & 26.49 & 28.16 & 26.84 & 26.04 & 26.5 & \gemmanine \\ 
        \textbf{Prompt 3} & 25.83 & 28.74 & 25.7 & 26.49 & 27.11 & 26.84 & 27.68 & 25.58 &  \\ 
        \textbf{} & ~ & ~ & ~ & ~ & ~ & ~ & ~ & ~ & ~ \\ 
        \textbf{Prompt 1} & 39.26 & 41.61 & 25.7 & 26.69 & 23.68 & 26.05 & 26.04 & 23.5 &  \\ 
        \textbf{Prompt 2} & 37.6 & 39.34 & 25.9 & 27.89 & 27.89 & 25 & 26.96 & 24.65 & \gemmats \\ 
        \textbf{Prompt 3} & 38.22 & 37.68 & 25.1 & 26.1 & 23.95 & 25.53 & 29.95 & 26.96 &  \\ 
        \textbf{} & ~ & ~ & ~ & ~ & ~ & ~ & ~ & ~ & ~ \\  
        \textbf{Prompt 1} & 28.1 & 29.81 & 27.09 & 27.09 & 27.37 & 26.58 & 25.35 & 25.35 &  \\ 
        \textbf{Prompt 2} & 26.03 & 27.95 & 28.09 & 24.9 & 27.11 & 27.63 & 25.35 & 25.12 & \llamas \\ 
        \textbf{Prompt 3} & 24.59 & 23.81 & 26.1 & 25.3 & 25.53 & 27.11 & 28.11 & 25.12 &  \\ 
        \textbf{} & ~ & ~ & ~ & ~ & ~ & ~ & ~ & ~ & ~ \\ 
        \textbf{Prompt 1} & 31.4 & 24.64 & 26.69 & 26.49 & 27.29 & 25.79 & 27.19 & 25.58 &  \\ 
        \textbf{Prompt 2} & 31.4 & 25.65 & 25.5 & 27.09 & 27.11 & 26.32 & 26.04 & 24.65 & \llamax \\ 
        \textbf{Prompt 3} & 27.48 & 28.16 & 26.29 & 26.49 & 27.11 & 25.79 & 26.04 & 25.12 &  \\ 
        \textbf{} & ~ & ~ & ~ & ~ & ~ & ~ & ~ & ~ & ~ \\ 
        \textbf{Prompt 1} & 51.03 & 53.00 & 43.03 & 27.29 & 48.16 & 32.11 & 30.41 & 30.18 &  \\ 
        \textbf{Prompt 2} & 53.72 & 57.35 & 41.24 & 29.28 & 50.00 & 32.63 & 30.41 & 30.65 & \aya \\ 
        \textbf{Prompt 3} & 48.14 & 55.28 & 41.63 & 29.88 & 51.05 & 33.42 & 74.19 & 85.25 &  \\ 
         \textbf{} & ~ & ~ & ~ & ~ & ~ & ~ & ~ & ~ & ~ \\ 
        \textbf{Prompt 1} &64.26 &66.87 &18.33 &59.56 &13.95 &13.95&0.23 &2.76  &  \\ 
        \textbf{Prompt 2} &62.81 &65.63 &49.00 &59.76 &41.05&3.68 &0.00 &0.69 & \gptf \\ 
        \textbf{Prompt 3} &6.2 &14.29 &14.34 &46.61 &19.47&5.00 &0.00 &0.00 &  \\ 
        \bottomrule
    \end{tabular}
    }
    \caption{Prompt experiments based on three distinct native proverbs.}
    \label{tab:nativeprompts}
\end{table*}

\clearpage

\section{Generation Results}
\label{app:generation-results}

In Table \ref{generation-scores} This table presents the performance evaluation metrics for Task 3, which involves generating proverbs. The table includes three key evaluation scores: ChrF, BLEU, and Translation Edit Rate (TER). These metrics are used to assess the quality and accuracy of the generated proverbs compared to the reference (ground truth) proverbs. Results show using BLEU score might be challenging for this tasks and ter doesn't tell us a clear picture of improvement.

\begin{table*}[!ht]
    \centering
    \resizebox{0.8\linewidth}{!}{%
    \begin{tabular}{l|ll|ll|ll|ll|l}
    \hline
         \multirow{2}{4em}{} & \multicolumn{2}{|c|}{\textbf{Amharic}}  & \multicolumn{2}{|c|}{\textbf{Afaan Oromoo}}& \multicolumn{2}{|c|}{\textbf{Tigrinya}} & \multicolumn{2}{|c|}{\textbf{Ge'ez}} & \textbf{English} \\ 
        \textbf{} & \textit{native} & \textit{english} & \textit{native} & \textit{english} & \textit{native} & \textit{english} & \textit{native} & \textit{english} & \\ \midrule
        \textbf{\llamathree} & ~ & ~ & ~ & ~ & ~ & ~ & ~ & ~ & ~ \\ 
        ChrF & 1.83 & 1.94 & 13.79 & 7.54 & 1.99 & 1.81 & 9.98 & 9.37 & 22.41 \\ 
        ter & 1295.16 & 388.73 & 578.43 & 654.78 & 1041.28 & 270.42 & 907.84 & 224.33 & 447.44 \\ 
        BLEU & 0.02 & 0.01 & 0.09 & 0.02 & 0.01 & 0.01 & 0.05 & 0.05 & 3.73 \\ \hline
        \textbf{\gemmanine} & ~ & ~ & ~ & ~ & ~ & ~ & ~ & ~ & ~ \\ 
        ChrF & 1.84 & 1.20 & 8.39 & 4.24 & 2.61 & 0.73 & 6.45 & 3.50 & 6.58 \\ 
        ter & 1371.83 & 1298.30 & 950.79 & 2054.45 & 629.45 & 1145.38 & 1804.11 & 2416.84 & 2423.57 \\ 
        BLEU & 0.02 & 0.00 & 0.04 & 0.00 & 0.02 & 0.00 & 0.02 & 0.00 & 0.77 \\ \hline
        \textbf{\gemmats} & ~ & ~ & ~ & ~ & ~ & ~ & ~ & ~ & ~ \\ 
        ChrF & 1.34 & 1.21 & 8.41 & 10.17 & 1.72 & 1.03 & 8.44 & 8.97 & 23.18 \\ 
        ter & 1015.48 & 900.26 & 459.48 & 435.41 & 862.70 & 753.32 & 417.14 & 258.83 & 528.82 \\ 
        BLEU & 0.01 & 0.00 & 0.03 & 0.01 & 0.01 & 0.00 & 0.01 & 0.04 & 4.75 \\ \hline
        \textbf{\llamas} & ~ & ~ & ~ & ~ & ~ & ~ & ~ & ~ & ~ \\ 
        ChrF & 2.23 & 2.74 & 10.12 & 5.73 & 3.72 & 3.03 & 11.30 & 6.80 & 21.61 \\ 
        ter & 628.61 & 354.65 & 946.16 & 1389.60 & 346.14 & 375.99 & 813.35 & 697.51 & 370.52 \\ 
        BLEU & 0.02 & 0.02 & 0.09 & 0.01 & 0.02 & 0.02 & 0.04 & 0.02 & 3.06 \\ \hline 
        \textbf{\llamax} & ~ & ~ & ~ & ~ & ~ & ~ & ~ & ~ & ~ \\ 
        ChrF & 5.29 & 4.90 & 18.11 & 10.16 & 3.38 & 2.54 & 9.73 & 9.06 & 31.25 \\ 
        ter & 179.81 & 157.10 & 165.60 & 332.72 & 310.93 & 191.25 & 157.21 & 146.29 & 106.14 \\ 
        BLEU & 0.06 & 0.05 & 0.24 & 0.05 & 0.03 & 0.01 & 0.04 & 0.04 & 13.68 \\ \hline
        \textbf{\aya} & ~ & ~ & ~ & ~ & ~ & ~ & ~ & ~ & ~ \\ 
        ChrF & 6.44 & 5.58 & 19.17 & 4.70 & 4.71 & 2.80 & 9.51 & 8.62 & 19.17 \\
        ter & 132.58 & 128.09 & 165.50 & 965.69 & 133.47 & 115.04 & 158.54 & 135.70 & 112.41 \\ 
        BLEU & 0.37 & 0.14 & 0.61 & 0.01 & 0.14 & 0.03 & 0.04 & 0.05 & 4.34 \\ \hline 
        \textbf{\gptf} & ~ & ~ & ~ & ~ & ~ & ~ & ~ & ~ & ~ \\
        ChrF & 5.63 & 0.03 & 16.94 & 3.27 & 6.38 & 4.70 & 6.00 & 3.88 & 50.00 \\ 
        ter & 132.58 & 128.09 & 165.50 & 965.69 & 133.47 & 115.04 & 191.74 & 291.64 & 112.41 \\
        BLEU & 0.37 & 0.14 & 0.61 & 0.01 & 0.14 & 0.03 & 0.05 & 0.02 & 4.34 \\ \bottomrule
    \end{tabular}
    }
    \caption{ChrF, Bleu, translation edit rate (ter) scores for Task 3: Proverb Generation Task}
    \label{generation-scores}
\end{table*}
\FloatBarrier

\end{document}